\RequirePackage{scrlfile}
\PreventPackageFromLoading{subfig}
\documentclass[letterpaper, 10 pt, conference]{ieeeconf}  

\IEEEoverridecommandlockouts                   
\usepackage{graphics} 
\usepackage{epsfig} 
\usepackage{float}
\usepackage{mathptmx} 
\usepackage{times} 
\usepackage{amsmath} 
\usepackage{amssymb}  
\usepackage{subcaption}
\usepackage{svg}
\usepackage{hyperref}
\usepackage{pgf,tikz}
\usepackage{mathrsfs}
\usetikzlibrary{arrows}
\usepackage{xcolor}
\usepackage{stackengine,scalerel}
\usepackage[pdf]{pstricks}
\usepackage[T1]{fontenc}
\usepackage[utf8]{inputenc}
\usepackage[none]{hyphenat}

\stackMath
\newcommand*\circled[1]{\tikz[baseline=(char.base)]{
   \node[shape=circle,draw,inner sep=0.2pt] (char) {#1};}}
   
\title{\LARGE \bf
Fusing Laser Scanner and Stereo Camera in Evidential Grid Maps
}

\author{Michelle Valente$^{1}$, Cyril Joly$^{1}$ and Arnaud de la Fortelle$^{1,2}$% <-this % stops a space
\thanks{*This work was supported by the International \emph{Chair Drive for All}}% <-this % stops a space
\thanks{$^{1}$Michelle Valente, Cyril Joly and Arnaud de la Fortelle are with Center of Robotics, Mines ParisTech, PSL Research University, 60 boulevard Saint-Michel, 75006 Paris, France. {\tt\small \{michelle.valente, cyril.joly, arnaud.de\_la\_fortelle\}@mines-paristech.fr}}%
\thanks{$^{2}$Arnaud de la Fortelle is also with California PATH, University of California, Berkeley, USA}%
}%

\begin{document}
% \twocolumn[{%
% \renewcommand\twocolumn[1][]{#1}%
\maketitle
% \begin{center}
%     \centering
%     \includegraphics[height=3.5cm]{images/teaser4.png}
%     \captionof{figure}{Test caption}
% \end{center}%
% }]

%%%%%%%%%%%%%%%%%%%%%%%%%%%%%%% Abstract %%%%%%%%%%%%%%%%%%%%%%%%%%%%%%%%%%%%%
\begin{abstract}

Automation driving techniques have seen tremendous progresses these last years, particularly due to a better perception of the environment. In order to provide safe yet not too conservative driving in complex urban environment, data fusion should not only consider redundant sensing to characterize the surrounding obstacles, but also be able to describe the uncertainties and errors beyond presence/absence (be it binary or probabilistic). This paper introduces an enriched representation of the world, more precisely of the potential existence of obstacles through an evidential grid map. A method to create this representation from 2 very different sensors, laser scanner and stereo camera, is presented along with algorithms for data fusion and temporal updates. This work allows a better handling of the dynamic aspects of the urban environment and a proper management of errors in order to create a more reliable map. We use the evidential framework based on the Dempster-Shafer theory to model the environment perception by the sensors. A new combination operator is proposed to merge the different sensor grids considering their distinct uncertainties. In addition, we introduce a new life-long layer with high level states that allows the maintenance of a global map of the entire vehicle's trajectory and distinguish between static and dynamic obstacles. Results on a real road dataset show that the environment mapping data can be improved by adding relevant information that could be missed without the proposed approach.

\end{abstract}
%%%%%%%%%%%%%%%%%%%%%%%%%%%%%%%%%%%%%%%%%%%%%%%%%%%%%%%%%%%%%%%%%%%%%%%%%%%%%%

%%%%%%%%%%%%%%%%%%%%%%%%%%%%%%% Introduction %%%%%%%%%%%%%%%%%%%%%%%%%%%%%%%%%
\section{Introduction}

Correctly understanding the environment is a crucial process for autonomous driving. For this purpose, the information coming from one or several sensors needs to be collected, analyzed and stored. Maps are commonly used as a tool to this end by storing different kinds of information, such as metric, semantic or topological. Similar to humans, an intelligent vehicle can use them to safely move from one location to another, and moreover, to discover where it is located inside the map: this is the base idea for SLAM and further refinements.

When it comes to exploiting the map data, a challenging problem is to get clear insurance about presence or absence of an object (called also obstacle though it can be a pedestrian): this is safety critical since a false negative may lead to a fatality, as recent crashes have shown. Now there is a large number of sensors available for intelligent vehicles to perceive the environment (e.g. laser scanners, sonars, stereo and mono cameras). Each sensor has its own particular limitations and advantages, in addition to different uncertainty characteristics. Due to these limitations, a single sensor cannot provide alone a robust reconstruction of the surroundings and hence cannot reliably perform important tasks like obstacle avoidance, path planning and localization. 

To enhance the accuracy of the environment perception, the data from multiple sensors must be merged and the result has to improve also the knowledge of presence or absence. This type of fusion can be performed at different levels of representation and by the use of different fusion methods. Occupancy grids \cite{grid} are often applied as a representation of the sensor data and its uncertainty, by mapping the obstacles detected by sensors. The mapping is performed by storing the probability of an area being occupied or free in each cell of the grid. There are several methods to merge sensor data into a grid map in order to deal with the uncertainty, such as Bayesian \cite{grid}, Fuzzy \cite{fuzzy} and Evidential \cite{evidentialgrid} methods.	

\begin{figure}
\centering
	\includegraphics[width=0.85\linewidth]{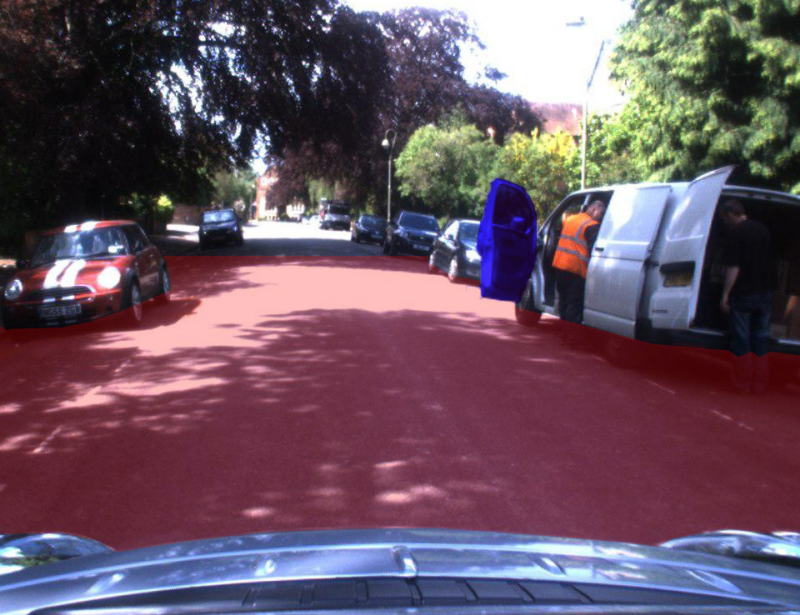}
	\caption{Example in which the sensor fusion added relevant information to the environment representation. In red is illustrated the laser scanner range and in blue an open door detected by the stereo camera but not by the laser scanner.}
	\label{fig:teaser}
\end{figure}

\begin{figure*}[h]
	\centering
	\includegraphics[scale=0.6]{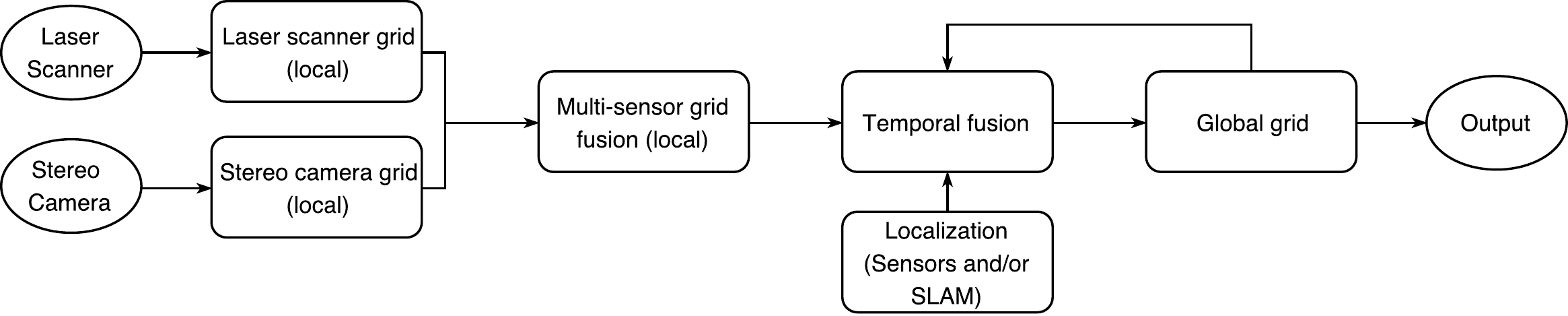}
	\caption{Schematics of the proposed system.}
	\label{fig:System}
\end{figure*}

In this work we have developed both a more robust representation of occupation (i.e. presence/absence) and an architecture that focuses on the data fusion process of different sensors to obtain this representation of the current and previous traversed driving environment. The proposed approach applies the Evidential framework to merge occupancy grids created by two different sensors: a 2D laser scanner and a stereo camera. We choose 2D laser scanners, instead of 3D laser scanners to address the demand of the automobile manufacturers to maintain a reasonable price and smaller size of the sensors for intelligent vehicles. First, we propose a novel evidential sensor model for the stereo-camera based on the disparity map and a Gaussian observation model. Next, we apply the state-of-the-art laser scanner evidential model \cite{bonnifait} and we introduce a new combination rule to merge the two different grids, taking into consideration the difference in confidence between the sensors. Sequentially, we apply a temporal grid fusion method to obtain a global map of the traversed path of the vehicle. Finally, we introduce a new life-long map layer based on the evidential beliefs that distinguishes between static, dynamic and not recently updated states. 

The remainder of the paper is organized as follows. First, we present the related work in Section~\ref{relatedwork.sec}; the algorithms are detailes in Section~\ref{evidentialgrid.sec}; experimental results are presented in Section~\ref{results.sec}; finally conclusion and perspectives are given in Section~\ref{conclusion.sec}.

%%%%%%%%%%%%%%%%%%%%%%%%%%%%%%%%%%%%%%%%%%%%%%%%%%%%%%%%%%%%%%%%%%%%%%%%%%%%%%
%%%%%%%%%%%%%%%%%%%%% Multi-sensor evidential grid mapping %%%%%%%%%%%%%%%%%%%
\section{Related Work}\label{relatedwork.sec}

In the literature, the most popular framework for mapping is the Bayesian; it represents the sensor's uncertainty by means of probability \cite{bayesian_occupancy}\cite{bayesian_occupancy2}\cite{bayesian_occupancy3}. On the other hand, the evidential approach, which is based on the Dempster-Shafer theory (DST) \cite{shafer}, has recently received attention because of its robust combination methods able to manage both noisy and conflicting information. DST allows conflict to be addressed directly, by modeling and differentiating lack of information from conflicting information. Moreover, it is a useful tool to combine different sensor information with distinct uncertainty parameters.  

Pagac et al. \cite{evidentialgrid} were the first authors to adapt the Bayesian sensor model to fit in the evidential framework using sonar data. Since then, there were several work on building evidential grid maps using different types of sensors, such as laser scanners \cite{laser_moras}\cite{laser_mobile}, radar \cite{radar} and stereo cameras \cite{stereo_evidential}. This different type of approach to occupancy grid maps brought new features like management of conflict data and combination operators for fusion.

In particular, evidential grid maps have recently received a lot of attention in the field of intelligent vehicles. Moras et al. \cite{bonnifait}\cite{bonnifait2}\cite{scene_understanding} presented their work on evidential grid maps to perceive the vehicle's environment and detect moving obstacles using a laser scanner. In \cite{scene_understanding} the grid map created by a laser scanner is enhanced by the fusion with a geo-referenced map. The authors perform temporal fusion to detect moving obstacles; however, they use a discounting factor to maintain only the recent map, which ultimately forgets and loses the previous mapped information about the environment. Trehard et al. \cite{fawzi}\cite{fawzi2} build as well an evidential grid map with a laser scanner using the same type of sensor model; but their focus is not on the current environment scene understanding, but rather on the localization of the vehicle performing SLAM. The use of stereo cameras were proposed as well in \cite{stereo_evidential} for evidential grid maps, where the authors introduce a sensor model based on the disparity map to directly obtain the free and occupied space. However, because of the high uncertainty of the mapping with a stereo camera, the results with only this sensor are not reliable to detect both free and occupied space and could not be directly applied to autonomous driving. 

Recent work shows the potential use of evidential grid maps for different tasks than only mapping, such as tracking of obstacles \cite{tracking_mapping}\cite{tracking2} and trajectory planning \cite{trajectory}\cite{journal_path}. To perform these tasks in a safe matter for an intelligent vehicle, the fusion of several perception sensors is essential. There are some studies in the literature where a 2D laser scanner and a camera are used together to perform navigation tasks. Lin et al. \cite{fusion_bayesian} proposed an approach for SLAM using a 2D laser scanner and 3D information from a stereo camera for indoor environments based on the Bayesian occupancy grids. In a similar way, Moghdan et al. \cite{fusion_bayesian2} explore the fusion of stereo camera and 2D laser scanner in Bayesian occupancy grids to perform path planning. In order to better handle the uncertainties of the sensors, this papers  presents a powerful framework based on evidential maps and fusion between such maps that provides more information about object presence 
and hence allows to better distinguish between static and mobile obstacles. To our knowledge, this is the first work presenting data fusion with different sensors in evidential maps and it shows it can be further developed with more than 2 sensors.
%%%%%%%%%%%%%%%%%%%%%%%%%%%%%%%%%%%%%%%%%%%%%%%%%%%%%%%%%%%%%%%%%%%%%%%%%%%%%%
%%%%%%%%%%%%%%%%%%%%% Multi-sensor evidential grid mapping %%%%%%%%%%%%%%%%%%%
\section{Multi-sensor evidential grid mapping}\label{evidentialgrid.sec}

The proposed approach consists in computing two evidential grid maps, one for the laser scanner and one for the stereo camera, and fuse them to have a common representation of the environment. Subsequently, we perform temporal fusion over the new grid and the stored grid to generate a global map. Finally, the output of the system is a global map that represents the environment where the vehicle drove. The architecture of the whole system is presented in Fig. \ref{fig:System}.

The creation of a grid is performed by interpreting the raw data of the sensors to metric information, which will represent the presence or absence of obstacles on that location. Moreover, to have a reliable map, we need to model the uncertainty of the sensor into the grid cells. We present a sensor model for a 2D laser scanner and for a stereo camera, that will deal with the sensor uncertainty to create the grid by using Dempster-Shafer theory of evidence \cite{shafer}. The evidential grid for the two sensor models has as a frame of discernment $\Omega=\{O,F\}$, referred as the states occupied (O) and free (F) of each cell. Therefore, the power set is defined as $2^\omega = \{ \emptyset, F, O, \Omega\}$, and each cell will store a basic belief assignment (BBA) with four beliefs $[m(F), m(O), m(\Omega), m(\emptyset)]$. Each belief represents respectively the evidence of being free, occupied, unknown or conflict.  

\subsection{Sensor model for laser scanner }

Laser data gives reliable information about free and occupied space in the environment. At each time step, the sensor provides a scan, which corresponds to a set of points measured during one laser rotation. Although this process is not instantaneous, we assume it is fast enough to map all points at the same time on the grid. We apply the solution proposed in \cite{bonnifait} to determine the evidence of a cell in a grid map by increasing the occupied evidence where there are laser impacts and the free evidence on the crossed cells. 

The proposed method defines the BBA of each cell in the laser scanner grid $LG$ at timestamp $t$ as follow:

\begin{equation}
\begin{split}
&\ \ \ \ \ m_{LG,t}(A) = \lambda \ \ \quad \ \ m_{LG,t}(B) = 0 \\  
&\ \ \ \ \ m_{LG,t}(\emptyset) = 0 \ \ \quad \ \ \ m_{LG,t}(\Omega) = 1 -\lambda \\
with \ A = &\begin{cases}
O \ if \ cell \ impacted \\
F \ if \ cell \ crossed
\end{cases}B = \begin{cases}
F \ if \ cell \ impacted \\
O \ if \ cell \ crossed
\end{cases}
\end{split}
\end{equation}

where $\lambda$ represents the laser scanner confidence. 

\subsection{Sensor model for stereo camera}

The second sensor used in our system is a binocular stereo-vision sensor. Our goal is not to estimate the free space with this sensor, but rather to improve the detection of obstacles that could be challenging for the laser scanner. Although 2D laser scanner are reliable and accurate, certain obstacles can be missed because of its height limitations, for example a door of a truck opened or a high gate. In addition, once one obstacle is detected it cannot detect anything further. Considering this, we define a stereo camera model that is based on the disparity space for obstacle detection. 

After receiving the two raw stereo images as input, we perform image rectification to bring the two images to a common image plane, which simplifies image matching methods. Once this is executed, the disparity map between the images is computed and the method for obstacle detection and mapping described below is performed. 

1) V-disparity map and ground estimation: the first step is to generate the V-disparity map by accumulating the pixels with the same disparity along the rows of the disparity image. As presented in \cite{vdisparity}, the V-disparity space can provides a representation of the geometric structure of the road and, for this reason, it can be used to estimate the ground plane pixels.

\begin{figure}
\centering
	\includegraphics[width=0.8\linewidth]{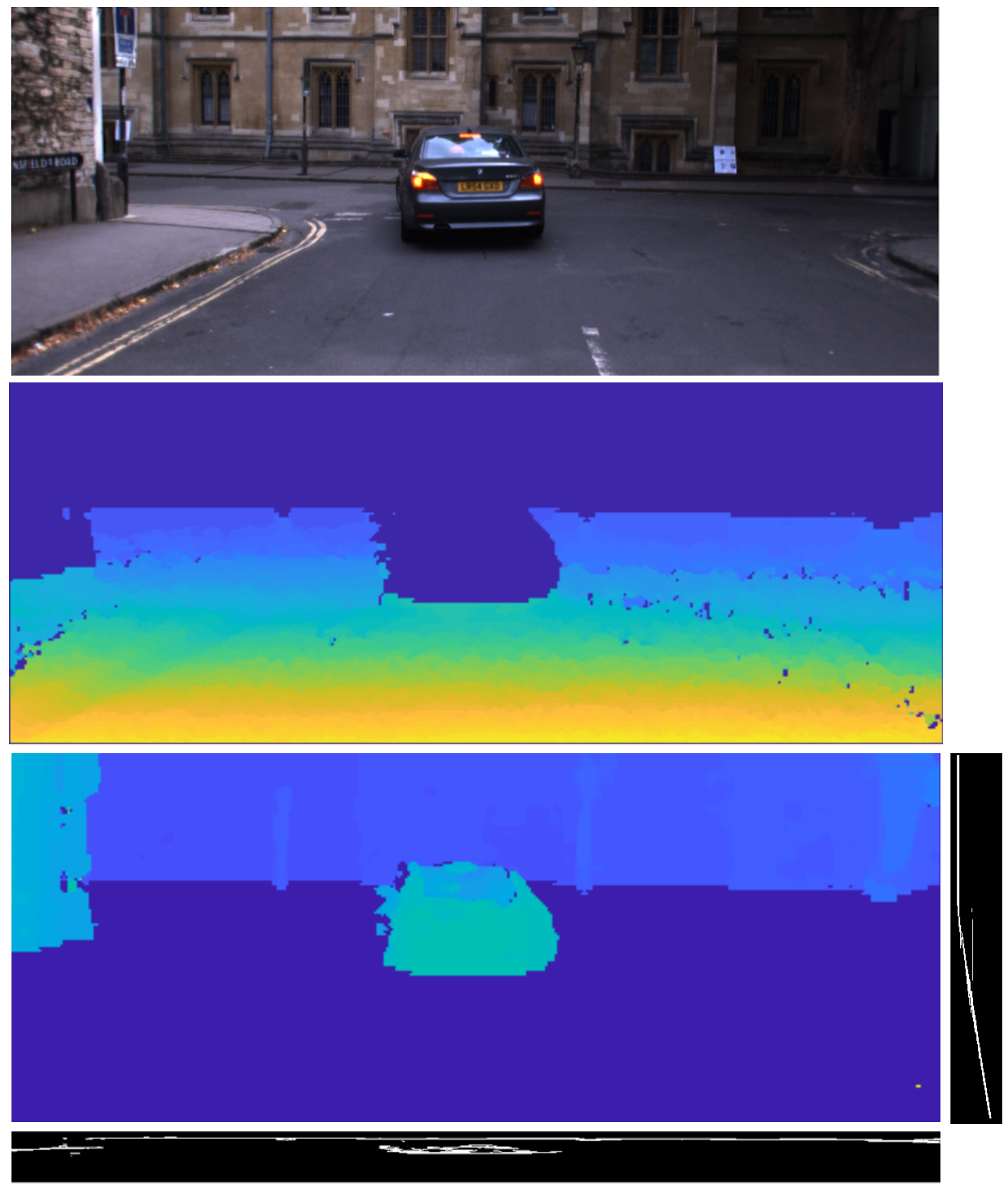}
	\caption{Top: original image from one of the cameras. Middle: ground disparity map. Bottom: obstacle disparity map with the V-disparity map on the right and the U-disparity map on bottom.}
	\label{fig:stereo}
\end{figure}

2) Obstacles detection: once we have defined the ground plane pixels, one can classify the obstacles pixels by separating the pixels that are over the ground and thresholding the height of the pixels. Sequentially, the obstacles pixels are mapped to the U-disparity space \cite{udisparity}. Instead of projecting the pixels related to the rows of the images, like in the V-disparity space, now we project them related to the columns. Fig. \ref{fig:stereo} shows the separation of the obstacles pixels along with the V-disparity and U-disparity maps.

3) Obstacle projection: the next step is to project the obstacles pixels to the world. First, we select on the U-disparity map the cells are over a threshold that eliminates false obstacle detections. Then, we adopt the method detailed in \cite{perrollaz_grid} to project these cells to the world. The stereo coordinate system has as origin the point $O_S$, which is the middle point of the baseline. The detection plane $P_D$ (supporting the grid) coordinate system has as origin the point $O_D$, the projection of $O_S$ on the plane. Considering $U_D$ the coordinates of a point in the U-disparity map $(u,d)$ and $X_D$ the 2D point related to the camera coordinate system $(x,y)$ on the detection plane, the transformation between $U_D$ and $X_D$ is performed by the function $G_D$:

\begin{equation}
\begin{split}
G_D : \mathbb{R}^2 \rightarrow \mathbb{R}^2 \\
U_D \mapsto X_D 
\end{split}
\end{equation}
with
\begin{equation}
\begin{cases}
&x = \frac{b}{2} + \frac{(u-u_0)\cdot b}{d}\\
&y = \frac{f \cdot b}{d}
\end{cases}
\end{equation}

where $b$ is the stereo camera baseline, ($u_0$, $v_0$) the principal point and $f$ the focal length.

\begin{figure}
\centering
\includegraphics[width=0.7\linewidth]{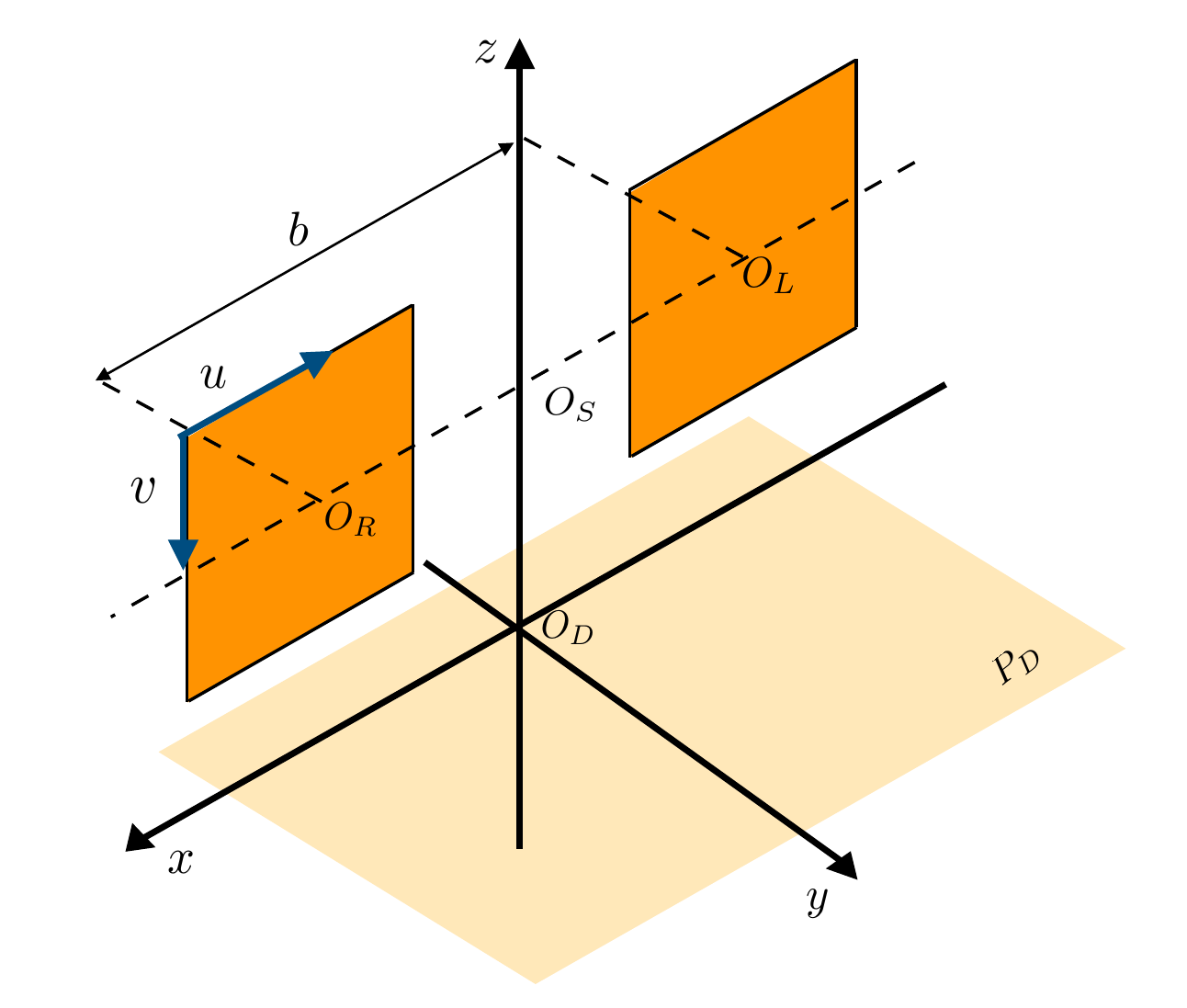}
\caption{Geometrical configuration of the stereo camera's coordinate system.}
\label{camera_coordinates}
\end{figure}

4) Evidential mapping: in the last step the points previously acquired are projected onto the 2D evidential occupancy grid. As a consequence of the pixel projection to the world, one pixel can have multiple observations in the grid. For this reason, we apply a Gaussian observation model centered on $U_D$; in addition, to transform directly the pixel to the grid cell, we linearize $G_D$ around the observed pixel. We base our model on the work of \cite{perrollaz_prob}, modifying it to consider only the $(u,d)$ coordinates. This modification allows to map the obstacle points straight from the U-disparity map to the occupancy grid, which can reduce the computation time of the transformation. Considering $U_i$ a point from the U-disparity map and $c_k$ a cell affected by this point, we define a projection factor $f(U_i,c_k)$. The factor is calculated under the Gaussian measurement model with linearization on the $G_d$ function as: 

\begin{equation}
f(U_i,c_k) \propto \exp(-\frac{1}{2} \|c_k -\mu_D^m \|^2_{K_D^m})
\end{equation}

where the Mahalanobis norm is defined by $\|v\|^2_A = v^T A^{-1} v$ and the mean and covariance are given by:

\begin{equation}
\begin{cases}
\mu_D^m = G_D(U_i)\\
K_D^m = J_G(U)\cdot\begin{bmatrix}
\sigma_u^2 & 0 \\
0 & \sigma_d^2
\end{bmatrix}\cdot J_G^T(U)
\end{cases}
\end{equation}

$J_G$ being the Jacobian matrix of the function $G_D$, while $\sigma_u$ and $\sigma_d$ are related to the errors produced by the stereo matching method. 

Since multiple obstacle points can be mapped to the same cell on the occupancy grid, we propose to apply an accumulation strategy to define the basic belief assignment of each cell. We define the total contribution of a cell as the sum of contributions of the $n$ points that are projected onto it:

\begin{equation}
C_{TOTAL}(c_k) = \sum\limits_{i=1}^n{f(U_i,c_k)\cdot I^{obst}_U(U_i) }
\end{equation}

where $I^{obst}_U(U_i)$ is the obstacle u-disparity map.

In order to avoid that taller objects have higher occupied evidence compared to shorter obstacles, we add an activation function based on the pixels contribution of each cell. In this way, the obstacle evidence grows quickly with respect to the number of obstacle pixels in that location with a threshold achieving the maximum stereo camera obstacle evidence. Considering this, the BBA of a cell $k$ in the stereo camera grid $SG$ at time slot $t$ is filled as follow:

\begin{equation}
\begin{split}
m_{SG,t}(F) =& 0 \\
m_{SG,t}(O) =& \tanh(\alpha_aC_{TOTAL}(c_k)) =  \\
&\frac{2}{1 + \exp(-2\alpha_aC_{TOTAL}(c_k))} - 1 \\
m_{SG,t}(\emptyset) =& 0 \\
m_{SG,t}(\Omega) =& 1 - \tanh(\alpha_aC_{TOTAL}(c_k))
\end{split}
\end{equation}

where $\alpha_a$ is a scale factor defined empirically. 

\subsection{Multi-sensor grid fusion}

\begin{figure}
\centering
\includegraphics[width=0.85\linewidth]{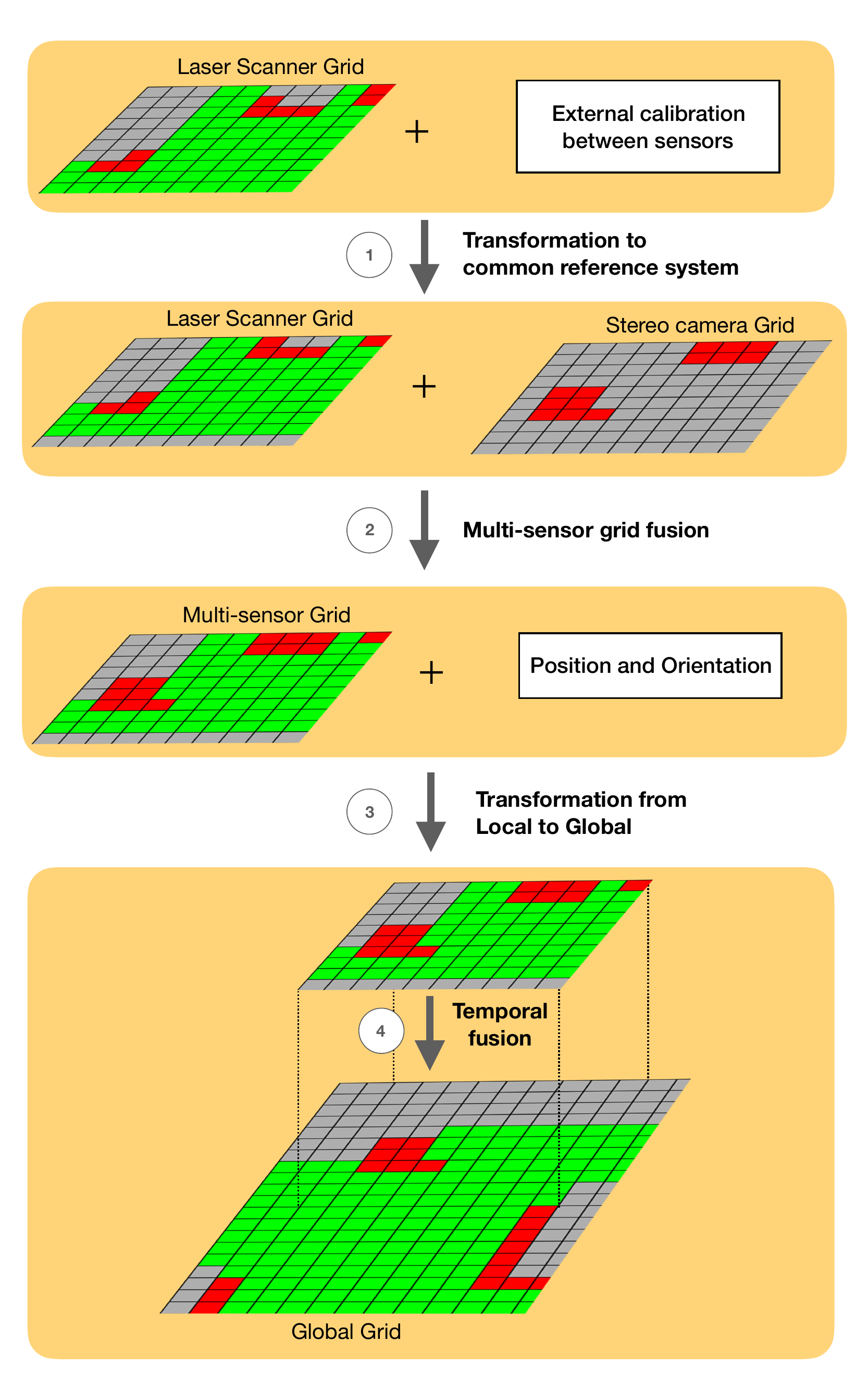}
\caption{Illustration of the multi-sensor and temporal fusion steps.}
\label{grid_transformations}
\end{figure}

To have a common representation of the environment, we create a combined map that merges the information from the two sensors grids. In order to perform the fusion, the grids need to be in the same coordinates. For this reason, each measurement has to be mapped to a joint reference system. This can be achieved using the results from the external calibration method between the stereo camera and the laser scanner. For this application, we consider that both of the sensors are already calibrated and these values are known. Fig. \ref{grid_transformations} first step illustrates this process. 

After we have obtained a common coordinate system, the fusion is executed in two steps. First, we apply a factor that models the \textit{a priori} knowledge related to the stereo camera sensor, in which the obstacle depth confidence decreases proportionally to the distance from the sensor. This reduces the influence of far obstacles detected by this sensor, and increases the influence of laser data in these cases. Considering this, a factor $\alpha_c$ is used to discount the probability masses of the stereo camera grid $SG$ at timestamp $t$ as follow:

\begin{equation}
m_{SG,t}^{\alpha_c}(A) = \begin{cases}
\alpha_c \cdot m_{SG,t}(A) \ \ \ \ \ \forall A \subsetneq \Omega \\
1 - \sum\limits_{B\subset \Omega} m_{SG,t}^{\alpha_c}(B) \ \ A = \Omega
\end{cases}
\end{equation}

where $\alpha_c$ is inversely proportional to the euclidean distance between the sensor location and the cell.

Sequentially, the laser scanner grid $LG$ and the stereo camera grid $SG$ are combined by applying the Dempster's rule (Eq. \ref{dempster}) generating a fused grid $FG$. We choose to use this combination rule because the two sources of information can be considered reliable, independent and have the same frame of discernment. It consists in two steps, first the conjunctive rule of combination denoted by $\circled{$\cap$}$ is applied (Eq. \ref{conjunctive}), and then the masses are normalized (Eq. \ref{normalize}). 

\begin{equation}
m_{FG,t} = m_{LG,t} \oplus m_{SG,t}^{\alpha_c}
\label{dempster}
\end{equation}

\begin{equation}
(m_1 \ \circled{$\cap$} \ m_2)(A)= \sum\limits_{A = B \cap C} m_1(B) \cdot m_2(C)
\label{conjunctive}
\end{equation}

\begin{equation}
m_1 \oplus m_2(A) = \begin{cases}
\frac{(m_1 \ \circled{$\cap$} \ m_2)(A)}{1 - (m_1 \ \circled{$\cap$} \ m_2)(\emptyset)} \ \ \ \ \ \forall A \subseteq \Omega \land A \neq \emptyset \\
0 \ \ \ \ \ \ \ \ \ \ \ \ \ \ \ \ \ \ \ \ \ \ A = \emptyset
\end{cases}
\label{normalize}
\end{equation}

\subsection{Temporal grid fusion}

After merging the two sensors grids, we need to accumulate the information gathered at different timestamps by applying a fusion process as illustrated by Fig. \ref{grid_transformations}. Like in the multi-sensor grid fusion, the map from the previously time slot and the new one need to be in the same coordinate system. For this purpose, the vehicle position and orientation at the current time slot have to be estimated. This can be performed by a Simultaneous Localization and Mapping (SLAM) method or by using information coming from sensors, such as GPS and IMU. This process is out of scope of this paper and we consider that this information is already provided. 

After the new local fused grid $FG$ is transformed to the global grid coordinates, we merge it to the previous global grid $GG$ by applying the Dempster's combination rule previously described:

\begin{equation}
m_{GG,t} = m_{GG,t-1} \oplus m_{FG,t}
\label{dempster_time}
\end{equation}

\subsection{Life-long grid}

Previous methods for evidential grid mapping \cite{bonnifait2}\cite{scene_understanding} choose to apply a time decay factor that decreases the belief of the cells during the temporal grid fusion. Over time, this factor erases mapped information and keeps only a recent map. In order not to lose the global map, we introduce a new life-long layer that provides more information than just the evidence of being free, occupied, uncertain or unknown. The new layer has five new different possible states: free space currently free (CF), free space currently unknown (CU), currently occupied space (CO), fixed occupied space (FO) and unknown (U). These states generate a semantical interpretation of the metric information provided by the evidential global grid. 

\begin{figure}
	\centering
	\includegraphics[scale=0.85]{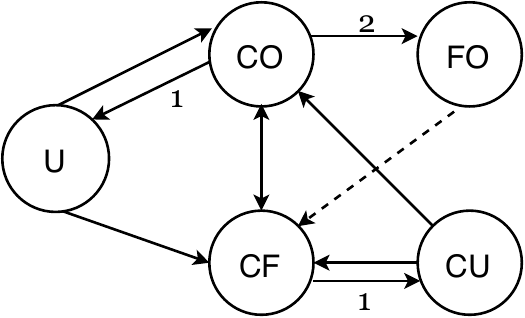}
	\caption{Diagram illustrating how the states of the life-long grid can change. 1 represents the timeout parameters and 2 the accumulator. The dashed lines represent a change in state that can happen in case a cell was wrongly classified as FO.}
	\label{fig:States}
\end{figure}

The current state of a cell is determined by analyzing the changes in the belief masses and by using two parameters: an accumulator and a timeout. First, every cell is assigned as unknown and when the evidence of free or occupied space increases they can be classified as CF or CO. After a number of time slots, the timeout parameter establishes if a CF cell state information is not recent anymore and should change to state CU. Another application of the timeout occurs when no new information arrives at a cell with state CO and the state needs to be changed to U. The second parameter, the accumulator, is used to determine if an obstacle can be considered fixed or not. It accumulates the number of time slots that this cell was classified as CO. After it arrives to a threshold, the cell state changes to FO. An important aspect is that the values assigned to the two parameters are dynamic according to the vehicle's velocity. The states behavior is summarized in Fig. \ref{fig:States}.

The introduced life-long layer permits us to obtain a global map with accurate information without loosing the previous mapped regions. In other words, the states of the  grid give us extra information that shows that a cell was not recently updated and it is not necessary to erase past information. Furthermore, the new layer defines the fixed obstacles in the map. The removal of dynamic information is crucial to localization algorithms, once they should only use fixed landmarks as reference. This knowledge can improve the localization coming from the sensors or even to allow the mobile robot to not rely on them anymore. 

%%%%%%%%%%%%%%%%%%%%%%%%%%%%%%%%%%%%%%%%%%%%%%%%%%%%%%%%%%%%%%%%%%%%%%%%%%%%%%

%%%%%%%%%%%%%%%%%%%%%%%%%%%%%%% Results %%%%%%%%%%%%%%%%%%%%%%%%%%%%%%%%%%%%%%
\section{Experimental Results}\label{results.sec}

The presented framework was tested on real-world urban traffic scenarios. We use different sequences from the public Oxford RobotCar dataset \cite{oxford}, which allows us to experiment the framework in different weather conditions and scenario situations. 
Our stereo camera sensor model uses the images coming from the wide baseline of the Point Grey Bumblebee XB3 (BBX3-13S2C-38) trinocular stereo camera. For the laser scanner sensor model we use one layer of the SICK LD-MRS 3D LIDAR. The laser scanner layer is extracted to simulate a low-cost 2D laser scanner, which is essential for the autonomous industry to maintain a reasonable price for the future intelligent vehicles. In order to perform the temporal fusion, the GPS/INS data is used in the estimation of the vehicle's position and orientation. The \href{https://youtu.be/SJUQO05Cu90}{video} \cite{video} presents the results at different steps of the proposed architecture.

\subsection{Multi-sensor local fusion results}

\begin{figure}
{
	\centering
    {
    \begin{subfigure}[t]{.155\textwidth}
        \includegraphics[width=1.05\linewidth]{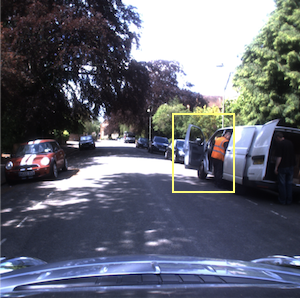}
    \end{subfigure}
    \begin{subfigure}[t]{.155\textwidth}
        \includegraphics[width=1.05\linewidth]{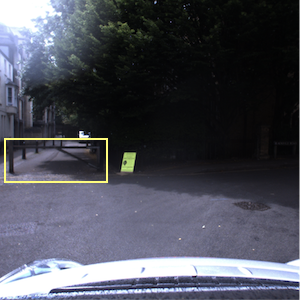}
       
    \end{subfigure}
    \begin{subfigure}[t]{.155\textwidth}
        \includegraphics[width=1.05\linewidth]{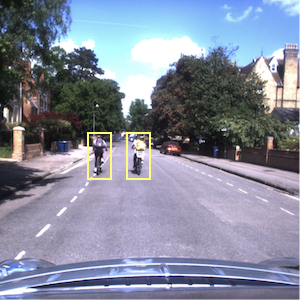}
    \end{subfigure}
    }
    \subcaption{Left camera image} 
    \par\medskip
    {
    \begin{subfigure}[t]{.155\textwidth}
        \includegraphics[width=1.05\linewidth]{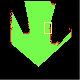}
    \end{subfigure}
    \begin{subfigure}[t]{.155\textwidth}
        \includegraphics[width=1.05\linewidth]{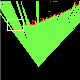}
    \end{subfigure}
    \begin{subfigure}[t]{.155\textwidth}
        \includegraphics[width=1.05\linewidth]{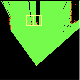}
    \end{subfigure}
    }
    \subcaption{Laser scanner grid}
    \par\medskip
    {
    \begin{subfigure}[t]{.155\textwidth}
        \includegraphics[width=1.05\linewidth]{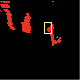}
    \end{subfigure}
    \begin{subfigure}[t]{.155\textwidth}
        \includegraphics[width=1.05\linewidth]{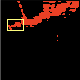}
    \end{subfigure}
    \begin{subfigure}[t]{.155\textwidth}
        \includegraphics[width=1.05\linewidth]{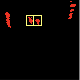}
    \end{subfigure}
    }
    \subcaption{Stereo camera grid}
    \par\medskip
    
    {
    \begin{subfigure}[t]{.155\textwidth}
        \includegraphics[width=1.05\linewidth]{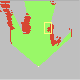}
    \end{subfigure}
    \begin{subfigure}[t]{.155\textwidth}
        \includegraphics[width=1.05\linewidth]{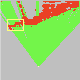}
    \end{subfigure}
    \begin{subfigure}[t]{.155\textwidth}
        \includegraphics[width=1.05\linewidth]{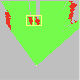}
    \end{subfigure}
    }
    \subcaption{Fusion grid}
}
    \caption{Examples of spatial multi-sensor grid fusion on three different time slots.}
    \label{spatial_fusion_result}
\end{figure}

The results for three different time slots are presented in Figure \ref{spatial_fusion_result}. The grid cell size was set to 0.25 m, which can model the environment considering the confidence of the two sensors. The occupied space in red represents the cells with $m(O)>0.5$, while the free space in green the cells with $m(F)>0.5$. 

Figure \ref{spatial_fusion_result} shows different situations where the fusion grid can provide a richer and more robust environment representation. As we can see by the camera image in the first situation, there is a vehicle with an open door which the laser scanner was not able to detect because of its height. However, in the fusion grid we have the door information due to the stereo camera mapping. The second example shows a similar problem, where a gate was not detected by the laser scanner because it was higher than the laser scanner, but was correctly added to the grid using the stereo camera information. The last example represents the situations where objects have few points detected by the laser scanner and can not be well mapped into the grid. In our example, we show that the two bikes in front of the vehicle are better recognized and mapped using the stereo camera images.

These results demonstrate how the proposed method is able to increase the environment understanding and thereby provide more information that can help different tasks of an intelligent vehicle. They illustrate how it is possible to improve the mapping process of a 2D laser scanner by fusing it with a stereo camera. The results also display how, even with less confidence in the data coming from the stereo camera, we can use it to add relevant obstacles to the grid. These improvements provide more knowledge about the current scene obstacles that can increase the safety of path planning methods.

\subsection{Temporal fusion results}

\begin{figure}
\centering
\includegraphics[scale=0.5]{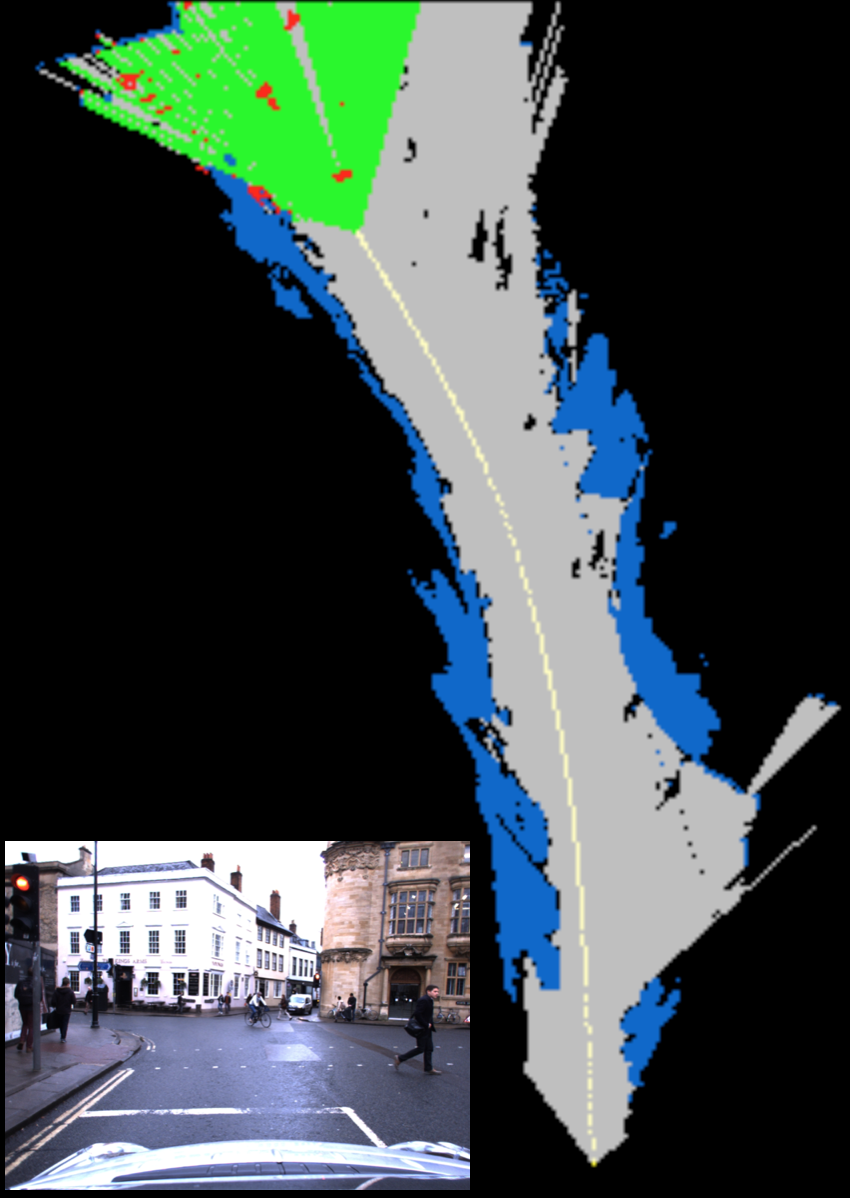}
\caption{Global map after performing temporal fusion. The colors represent the different states of the life-long grid in II-E.}
\label{temporal}
\end{figure}

The result from a sequence of time slots is presented in Figure \ref{temporal}. 
It presents the output of the life-long grid, where each state has a different color on the grid. The traversed region is represented by the color gray for the state free space currently unknown (CU) and the color blue for the fixed occupied space (FO). While the current scene is represented by the colors green for the free space currently free (CF) and red for the currently occupied space (CO). In addition, the color black represents the unknown (U) space and the yellow the trajectory of the vehicle during the data collection. 

The camera image is also presented in the Figure \ref{temporal} and it allows us to understand better what is currently being mapped. We can observe that the pedestrians and the bikes were correctly defined with the CO state and the light pole and buildings as FO. We can also note that some regions that were traversed by the vehicle ended with the state U. There are three possible reasons for this result: the sensors did not detect them, the fixed occupied threshold was not achieved or the last information was a moving obstacle in that region.

The global map created by the temporal fusion provides a robust estimation of the fixed obstacles and the previous free space. The fixed obstacles are fundamental for localization methods and by our method we can increase them by the spatial fusion and improve their confidence on being truly fixed by the life-long layer. Moreover, path planning methods can use the cells with states CU as an estimation of where it can be possible for the vehicle to move once it returns to a previously mapped location. 

\subsection{Quantitative evaluation}

In \cite{yager} Yager introduced the concepts of entropy and specificity in the framework of Dempster-Shafer's theory. These parameters are complementary and can be used to indicate the quality of evidence. They were applied in different papers \cite{uncertainty_evidential_grid}\cite{stereo_evidential} in order to obtain quantitative parameters to analyze the performance of evidential grid maps.

A high value of entropy can indicate inconsistency in the distribution of the mass beliefs. The entropy of a cell can be calculated as follow:

\begin{figure}[H]
\centering
\includegraphics[width=1.0\linewidth]{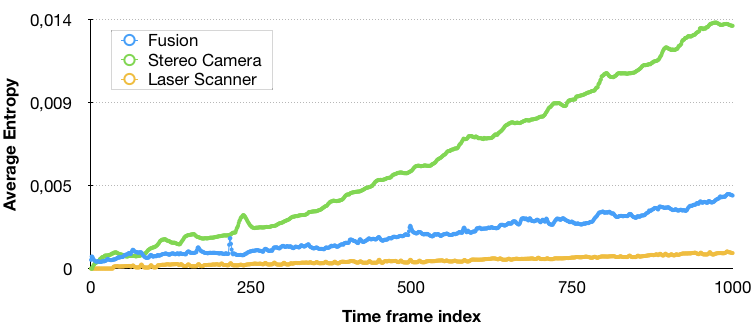}
\caption{Average entropy for each frame}
\label{graph_entropy}
\end{figure}

\begin{figure}[H]
\centering
\includegraphics[width=1.0\linewidth]{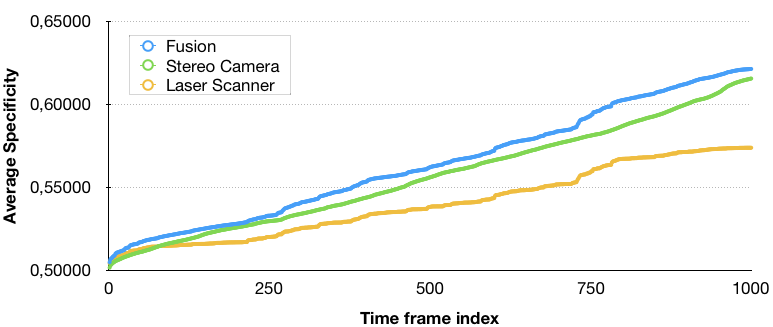}
\caption{Average specificity for each frame}
\label{graph_specificity}
\end{figure}

\begin{equation}
E_c = - \sum\limits_{A\subseteq\Omega}m(A)\cdot \ln(pl(A))
\end{equation}

where $pl(A)$ is the plausibility of $A$ (please refer to \cite{bonnifait2} for further details).

The specificity parameter provides an indication of how the belief mass is dispersed. Therefore, it has a higher value if the mass distribution is less doubtful. The specificity value of a cell can be calculated as follow: 

\begin{equation}
S_c = \sum\limits_{A\subseteq\Omega, A\neq\emptyset}\frac{m(A)}{card(A)}
\end{equation}

Considering the two parameters, we can conclude that: the lower the entropy, the more consistent is the evidence; and the higher the specificity less diverse it is. Therefore, for better certainty we need low entropy and high specificity. 

In Figure \ref{graph_entropy} and Figure \ref{graph_specificity} we present the average entropy and specificity for the global grid at each time frame. We compare our proposed method with the related work of laser scanner \cite{bonnifait} and stereo camera \cite{stereo_evidential} evidential grid built separately without performing sensor fusion.

In general, the multi-sensor fusion grid has higher specificity compared to the other two grids. Therefore, we can suppose that the grid created by the proposed method is less doubtful and has a more reliable representation of the environment. At the same time, the fusion grid has higher entropy than the laser scanner grid. This is expected since we are performing the fusion of two sources of information with different fields of view.  Nonetheless, it still remains a very low value of entropy compared to the stereo camera alone. One can conclude from these values that even with the small addition of entropy, it is still significant the addition of specificity that provides more relevant information about the environment and increase the quality of the map.

%%%%%%%%%%%%%%%%%%%%%%%%%%%%%%%%%%%%%%%%%%%%%%%%%%%%%%%%%%%%%%%%%%%%%%%%%%%%%%

%%%%%%%%%%%%%%%%%%%%%%%%%%%%%%% Conclusion %%%%%%%%%%%%%%%%%%%%%%%%%%%%%%%%%%%
\section{Conclusion and Future Work}\label{conclusion.sec}

In this paper we presented a framework to perform environment mapping by executing multi-sensor data fusion. We also designed a life-long layer that allows us to create global maps and distinguish the different types of information in the map. Results demonstrate how the fusion of the stereo camera grid with the laser scanner grid can enhance the quality of the mapping process and therefore better characterize the environment. Moreover, quantitative results show that, even with the fusion of different sources of information, we can still have a satisfactory evidential grid quality.

Future work will concentrate on improvements to the life-long layer and in the application of the proposed approach to increase the quality of other intelligent vehicle's tasks. First, a more complete analysis of the moving objects can be performed in order to better detect and track them. This can improve the certainty of the states by classifying the fixed and dynamic objects with a better precision. Additionally, the knowledge from the camera images can be used to complement the metric information mapped on this work by the application of semantical information acquired from deep learning methods. The semantical data can increase the number of states of the new layer and give more specific classifications about what is being mapped. 

\addtolength{\textheight}{-12cm}  

%%%%%%%%%%%%%%%%%%%%%%%%%%%%%%%%%%%%%%%%%%%%%%%%%%%%%%%%%%%%%%%%%%%%%%%%%%%%%%

%%%%%%%%%%%%%%%%%%%%%%%%%%%%%%%% Bibliography %%%%%%%%%%%%%%%%%%%%%%%%%%%%%%%%

%%%%%%%%%%%%%%%%%%%%%%%%%%%%%%%%%%%%%%%%%%%%%%%%%%%%%%%%%%%%%%%%%%%%%%%%%%%%%%

\end{document}